\documentclass{article}
\usepackage{arxiv}

\usepackage{graphicx}
\usepackage{subfig}
\graphicspath{./}
\usepackage{amsmath,amssymb,amsthm}
\usepackage{bm}
\usepackage{color}
\usepackage[normalem]{ulem}

\newlength{\figwidth}
\newlength{\figwidthtwo}
\newlength{\figwidththree}
\newcommand{\fref}[1]{Fig.\,\ref{#1}}
\newcommand{\tref}[1]{Table\,\ref{#1}}
\newcommand{\eref}[1]{Eq.\,(\ref{#1})}

\newcommand{\sref}[1]{Sec.\!~\ref{#1}}

\newcommand{\cref}[1]{Ref.\,\cite{#1}}

\newcommand\pN{\mathcal{N}}
\newcommand{\optbeta}{\beta^*}

\newcommand{\kld}{D_{\text{KL}}\infdivx}

\newcommand{\Nc}{\mathcal{N}}

\newcommand{\Rbb}{\mathbb{R}}

\newcommand{\mub}{{\boldsymbol{\mu}}}

\newcommand{\thetab}{{\boldsymbol{\theta}}}

\newcommand{\Sigmab}{\boldsymbol{\Sigma}}

\newcommand{\Ab}{\mathbf{A}}

\newcommand{\Db}{\mathbf{D}}

\newcommand{\Rb}{\mathbf{R}}

\newcommand{\Yb}{\mathbf{Y}}

\newcommand{\argmin}{\operatorname{argmin}}

\newcommand{\tr}{\operatorname{tr}}

\newcommand{\expectation}{\mathbb{E}}

\DeclareMathOperator*{\argmax}{arg\,max}

\renewcommand{\d}[1]{\ensuremath{\operatorname{d}\!{#1}}}
\renewcommand{\det}[1]{\operatorname{det}( #1 )}

\usepackage{amsmath,bm,amsfonts,amssymb,mathrsfs,mathtools}
\usepackage{xcolor}
\usepackage{booktabs}

\DeclarePairedDelimiterX{\infdivx}[2]{(}{)}{%
#1\;\delimsize\|\;#2%
}

\renewcommand{\headeright}{}
\renewcommand{\shorttitle}{Novel Probabilistic Transfer Learning}

\begin{document}

\title{Enhancing Polynomial Chaos Expansion Based Surrogate Modeling using a Novel Probabilistic Transfer Learning Strategy}
\author{Wyatt Bridgman\\
Sandia National Laboratories,\\
Livermore, California 95391, USA
\And
Uma Balakrishnan\\
Sandia National Laboratories, \\
Livermore, California 95391, USA
\And
Reese Jones\\
Sandia National Laboratories, \\
Livermore, California 95391, USA
\And
Jiefu Chen\\
University of Houston, \\
Houston, Texas, 77081, USA
\And
Xuqing Wu\\
University of Houston, \\
Houston, Texas, 77081, USA
\And
Cosmin Safta\\
Sandia National Laboratories, \\
Livermore, California 95391, USA
\And
Yueqin Huang\\
University of Houston, \\
Houston, Texas, 77081, USA
\And
Mohammad Khalil\thanks{corresponding:mkhalil@sandia.gov}\\
Sandia National Laboratories, \\
Livermore, California 95391, USA
}

\maketitle

\begin{abstract}

In the field of surrogate modeling, polynomial chaos expansion (PCE) allows practitioners to construct inexpensive yet accurate surrogates to be used in place of the expensive forward model simulations.
For black-box simulations, non-intrusive PCE allows the construction of these surrogates using a set of simulation response evaluations.
In this context, the PCE coefficients can be obtained using linear regression, which is also known as point collocation or stochastic response surfaces.
Regression exhibits better scalability and can handle noisy function evaluations in contrast to other non-intrusive approaches, such as projection.
However, since over-sampling is generally advisable for the linear regression approach, the simulation requirements become prohibitive for expensive forward models.
We propose to leverage transfer learning whereby knowledge gained through similar PCE surrogate construction tasks (source domains) is transferred to a new surrogate-construction task (target domain) which has a limited number of forward model simulations (training data).
The proposed transfer learning strategy determines how much, if any, information to transfer using new techniques inspired by Bayesian modeling and data assimilation.
The strategy is scrutinized using numerical investigations and applied to an engineering problem from the oil and gas industry.
\end{abstract}

\keywords{Transfer Learning, Polynomial Chaos Surrogates, Bayesian calibration, Subsurface characterization}

\section{Introduction}

Obtaining cheap, accurate surrogates for complex physics models is paramount in addressing a broad spectrum of research challenges. For example, they can illuminate high-dimensional relationships between input parameters and model responses~\cite{forrester2008}, which  makes them invaluable for model calibration \cite{von2011bayesian,dose2003bayesian}, uncertainty quantification (UQ), optimization~\cite{queipo2005surrogate}, and decision-making under uncertainty.
Hence, surrogates serve as a powerful toolset, enabling a deeper understanding of intricate system behaviors and paving the way for robust and informed decision-making in the face of uncertainty~\cite{jones1998efficient,williams2006gaussian,simpson2001kriging}.

A wide variety of approaches for surrogate modeling exist and span intrusive, physics-based methods, like reduced-order models, to data-driven neural network surrogates \cite{li2023surrogate,cai2021physics}.
In the context of surrogate modeling, polynomial chaos expansions (PCEs) \cite{wiener1938homogeneous,xiu2002wiener,ghanemstochastic} allow practitioners to construct inexpensive yet accurate surrogates to be used in place of the expensive forward model simulations.
PCEs originated as an approach to uncertainty quantification (UQ) based on multidimensional polynomials that are orthogonal with respect to the joint probability distribution of the input variables.
This includes uniform (Legendre polynomials) and Gaussian (Hermite polynomials)~\cite{ernst2012convergence} distributions.
In addition to UQ, PCEs can also be used to build effective surrogates for models by assuming a uniform density on parameters of interest over a domain~\cite{liu2001monte}.
Since the basis is complete, the coefficients of the polynomial expansion fully characterize the model and can be obtained using Galerkin projection~\cite{marzouk2007stochastic} or regression~\cite{blatman2011adaptive}.
PCEs have seen wide-spread application in diverse fields such as engineering, physics~\cite{xiu2003modeling,najm2009uncertainty}, finance and epidemiology~\cite{zhu2023stochastic}.
For black-box simulations, non-intrusive PCE allows the construction of surrogates using a set of simulation response evaluations.
In this context, the PCE coefficients can be obtained using linear regression, which is also known as point collocation or stochastic response surfaces.
Regression exhibits better scalability and can handle noisy function evaluations in contrast to other non-intrusive approaches, such as projection.
However, since over-sampling is generally advisable for the linear regression approach, the simulation requirements become prohibitive for expensive forward models.
We propose to leverage transfer learning ~\cite{pan2009survey,zhuang2020comprehensive}, whereby knowledge gained through similar PCE surrogate construction tasks (source domains) is transferred to a new surrogate-construction task (target domain) which has a limited number of forward model simulations (training data).

Transfer learning \cite{zhuang2020comprehensive,pan2009survey} has had an increasing number of applications in machine learning (ML) \cite{wang2018deep,tan2018survey} where data-hungry, many-parameter models are trained to capture abstract patterns in large datasets.
Two important categories of transfer learning are: \emph{feature-based} transfer and \emph{parameter-based} transfer.
Feature-based transfer assumes similarity can be found between the source and target features using approaches based on augmentation \cite{daume2009frustratingly}, mapping \cite{pan2010domain}, and clustering \cite{dai2007co}.
In parameter-based transfer, the focus is instead on exploiting similarity between parameters defining the source and target task models~\cite{lawrence2004learning,zhuang2015supervised,zhuang2017supervised} with methods like parameter sharing and regularization.
Here, we focus on parameter-based transfer since the assumption that the input-to-output map is similar between tasks has a variety of applications in physics and engineering, as we will illustrate in \sref{sec:subsurface}.

Parameter-based transfer learning techniques are often deterministic in that they yield a point estimate for the target task model parameters and so do not offer a strategy to quantify uncertainty.
In this work, we propose a probabilistic, parameter-based transfer learning strategy to address this gap.
The approach consists of transferring knowledge between Bayesian inverse problems where the source knowledge is represented as a prior over the target task which is informed by the posterior over the source task.
A key novelty of this methodology is controlling the transfer of by tempering the source posterior to avoid \textit{negative transfer}, whereby accuracy is impeded by the source information.
This is carried out by posing an optimization problem to determine the optimal amount of knowledge to carry over from the source task.

The remainder of the paper is organized as follows.
The next section, \sref{sec:methodology}, describes the probabilistic transfer learning methodology where an optimization and tempering procedure are introduced to control knowledge transfer.
Then \sref{sec:numerical_results} presents the result of three demonstrations.
First, domain adaptation of simple polynomial models is described in \sref{sec:domain_adaptation} and then extended to a more complex model in \sref{sec:ishigami}.
The proposed workflow is then applied to a subsurface scattering physics surrogate in \sref{sec:subsurface}.
Finally, a summary and a perspective on future work is given in \sref{sec:conclusion}.

\section{Methodology} \label{sec:methodology}
In this section, we describe the main methodological concepts used in the proposed transfer learning framework.
We first describe surrogate modeling and calibration with PCEs using standard Bayesian linear regression.
Next, transfer learning in the context of Bayesian inverse problems is introduced along with the key concept of tempering to control the amount of knowledge transferred from source to target.
In this work tempering takes the form of parametric transformations applied to distributions to diffuse probability mass and thereby control how much knowledge is transferred.
Different approaches to determine optimal tempering parameters are discussed, and several choices for the objective function motivated and described in detail.

\subsection{PCE Surrogate Modeling}

Polynomial Chaos expansions have a long history in computational science and engineering~\cite{wiener1938homogeneous,ghanem1991stochastic,xiu2002wiener,soize2004physical,crestaux2009polynomial,hadigol2018least}; here we review the fundamentals in the context of surrogate modeling.

Assume that a quantity of interest, $y$, of a physical system is modeled by
\begin{equation}
y = \mathcal{M} \left( \mathbf{x} \right) \ ,
\end{equation}
in which $\mathbf{x} = \left\{ x_1, \hdots, x_n \right\} \in \mathbb{R}^n$ is a vector of model parameters.
Assuming, without loss of generality, that $x_1, \hdots, x_n$, through some (usually linear) transformation of the corresponding model parameters vary within the $n$-dimensional hypercube $\left [ -1, 1 \right]^n$, one can construct a polynomial approximation of $y$, as
\begin{align}
y  = P\left( \mathbf{x} \right)
& = \theta_0 + \sum_{j=1}^n \theta_j \psi_j\left( x_j \right)
+  \sum_{j=1}^n \sum_{k=1}^j \theta_{jk} \psi_{jk}\left( x_j, x_k \right) \nonumber \\
& \ \ +  \sum_{j=1}^n \sum_{k=1}^j \sum_{h=1}^k  \theta_{jkh} \psi_{jkh}\left( x_j, x_k, x_h \right) + \cdots \ ,
\end{align}
where $\theta_{i\ldots}$ are coefficients and $\psi_{i\ldots}$ are, in the case of polynomial chaos expansions (PCE), multivariate polynomial basis functions.
The PCE basis is orthogonal with respect to the joint uniform probability density function (PDF) of $\mathbf{x}$ on $\left [ -1, 1 \right]^n$.
In practice, this PCE is truncated at a finite number of basis functions.
For a total order truncation, the number of terms, $N_{\textrm{PCE}}$, in the expansion is given by
\begin{equation}
N_{\textrm{PCE}} = \frac{\left( n+d \right) ! }{n ! \ d !} \ ,
\end{equation}
where $d$ is the degree of the expansion.
The coefficient vector, $\thetab  \in \mathbb{R}^{N_{\textrm{PCE}}}$, can be determined via linear regression using a training data set:
\begin{align}
X & = \left\{ \mathbf{x}^1, \hdots, \mathbf{x}^{N_p} \right\}  \\
Y & = \left\{ y^1, \hdots, y^{N_p} \right\}  \ ,
\end{align}
Generally, it is advisable to have $N_p > N_{\textrm{PCE}}$, resulting in an over-determined system of equations for $\thetab$.
Since the truncated PCE, $P_d \left( x \right)$, is an approximations of $y$, we will model each sample of  the data as
\begin{equation}
y^k = P_d \left( \mathbf{x}^k, \thetab \right) + \epsilon^k\ ,
\end{equation}
and assume that the error term $\epsilon^k$ is independently and identically distributed white Gaussian noise, i.e. $\epsilon^k \sim \pN(0, \gamma^2)$. In this case, the likelihood function for the coefficients, $\thetab$, has the form:
\begin{equation}\label{eq:likelihood}
p \left( D \vert \thetab \right) = \pN(\bar{\thetab}_{\rm lik}, \Sigmab_{\rm lik} ) \ ,
\end{equation}
with $D = \left\{ X, Y \right\}$,
\begin{align}
\bar{\thetab}_{\rm lik} & = \left( \Ab^T \Ab \right)^{-1} \Ab^T \, \Yb \\
\Sigmab_{\rm lik} & = \gamma^2 \left( \Ab^T \Ab\right)^{-1}  \ ,
\end{align}
and $\Ab$ is the so-called Vandermonde matrix of polynomial basis functions evaluated at the set of parameter vector realizations $X$.

Assuming, for reasons of conjugacy, that the prior PDF of $\thetab$ is also Gaussian,
\begin{equation}
p \left( \thetab \right) = \pN(\bar{\thetab}_{\rm pr}, \Sigmab_{\rm pr} ) \ ,
\end{equation}
the posterior PDF, given by Bayes' law, is also Gaussian of the form:
\begin{align}
p \left( \thetab \vert D \right) & = \frac{p \left( D \vert \thetab \right) \ p \left( \thetab \right) }{p \left( D \right) } \nonumber \\
& = \pN(\bar{\thetab}_{\rm post}, \Sigmab_{\rm post} ) \ ,
\end{align}
with mean $\bar{\thetab}_{\rm post}$ and covariance $\Sigmab_{\rm post}$:
\begin{align}
\bar{\thetab}_{\rm post} & = \left( \left( \Sigmab_{\rm pr} \right)^{-1} + \left( \Sigmab_{\rm lik} \right)^{-1} \right)^{-1} \left( \left( \Sigmab_{\rm pr} \right)^{-1} \bar{\thetab}_{\rm pr} + \left( \Sigmab_{\rm lik} \right)^{-1} \bar{\thetab}_{\rm lik} \right) \label{eq:mu_gaussian_mix}\\
\Sigmab_{\rm post} & = \left( \left( \Sigmab_{\rm pr} \right)^{-1} + \left( \Sigmab_{\rm lik} \right)^{-1} \right)^{-1}   \ . \label{eq:sig_gaussian_mix}
\end{align}

\subsection{Transfer Learning} \label{sec:transfer_learning}

Limited available data can be used to {\it partially} inform the PCE coefficients through the analysis of the likelihood function in \eref{eq:likelihood}.
Bayes' law, as shown in the previous section, provides the mechanism of further informing the PCE coefficients using knowledge gained prior to the analysis of the target data.
In a transfer learning setting, this prior knowledge is gained through the training of the PCE surrogate on other (possibly similar) data sets.
However, the Bayes' update \emph{should not} be applied as is in this scenario since the level of similarity of those other sets of data relative to the current one is not known {\it a priori}.

In this case, we propose the following extension to the Bayes' update to merge current knowledge gained, as captured by the likelihood function, and prior knowledge as encoded in the prior PDF:
\begin{equation}\label{eq:mod_bayes}
p \left( \thetab \vert D, \beta \right) \propto p \left( D \vert \thetab \right) \ p \left( \thetab \right)^\beta \ ,
\end{equation}
The {\it tempering} parameter $0 \leq \beta \leq 1$ allows the formalism to {\it diffuse} the prior knowledge, with the limiting cases of $\beta = 0$ eliciting no use of prior knowledge and  $\beta = 1$ reverting to the traditional Bayes' law.
This specific form of tempering via exponentiation diffuses bell-shaped priors by widening their extent, thus making the transferred information less certain;
it is related to the \emph{power prior} \cite{ibrahim2003optimality,ibrahim2015power} employed outside of transfer learning.
This added flexibility allows us to control how much knowledge in the PCE coefficients is transferred from the prior PDF to the posterior, with diminishing effect as $\beta$ approaches 0.

In this formalism, the single tempering parameter $\beta$ controls the amount of knowledge to be transferred from the source training task, with data $D_S$, to the target training task involving the construction of PCE surrogates, with target data $D_T$.
The target data $D_T$ is used in formulating the likelihood function, and the source data $D_S$ is used in a separate calibration task to provide a prior PDF on $\thetab$.

We can determine the tempering parameter $\beta$ using one of the following approaches:
\begin{itemize}
\item \emph{Hierarchical Bayes}: One can proceed with the joint inference of the PCE coefficients $\thetab$  and tempering parameter $\beta$ by formulating the joint posterior PDF:
\begin{align}
p \left( \thetab , \beta \vert D \right) & \propto p \left( \thetab \vert D, \beta \right) \ p \left( \beta \right)   \ ,
\end{align}
with $p \left( \beta \right)$ being the prior PDF on the tempering parameter $\beta$, for example a uniform PDF on [0,1].
\item \emph{Empirical Bayes}: Utilize a point estimate for $\beta$ that may be obtained through a separate optimization problem.
The associated objective function may be based on cross-validation or maximum expected data-fit, for example.
\end{itemize}

For this investigation, we will focus on the empirical Bayesian approach and formulate an optimization problem to obtain a point estimate $\optbeta$ for $\beta$, the tempering parameter.

\subsection{Tempering} \label{sec:tempering}

\emph{Tempering} is, in general, a means of reducing the certainty of previous information in the form of a mollified prior.
This usually involves changing the covariance structure of the target prior (source posterior) in a parametric way, such as convolving it with a kernel.
Here we use power tempering and focus on Gaussian priors.

To define tempering, let $\pi_S(\thetab | \beta)$ denote the likelihood (or posterior) over parameters $\thetab$ for the source task.
Here, $\pi_S$ is also a function of the tempering parameter $\beta$.
This provides a compact notation for indicating that a $\beta$-dependent tempering transformation is applied to the original, unmodified source likelihood $\pi_S(\thetab)$ to produce a tempered distribution $\pi_S(\thetab | \beta)$.
By taking the tempered source likelihood to be the prior for the target calibration, the posterior distribution for the target task is $\beta$-dependent and given by
\begin{equation}
\pi_p(\thetab| \beta)  \propto	\pi_S(\thetab| \beta)  	\pi_T(\thetab)
\end{equation}
where $\pi_T(\thetab)$ is the target likelihood.

In this work, we use focus on power tempering, a transformation defined by raising the distribution to a power $\beta$
\begin{equation}
\pi_S(\thetab| \beta) \propto \pi_S(\thetab)^\beta
\end{equation}
In the case of Gaussian distributions, the tempered source likelihood has the form $\pi_S( \thetab \vert \beta) = \Nc(\thetab; \mub_S,(1/\beta)\Sigmab_S)$, as raising a Gaussian to a power $\beta$ simply scales the covariance matrix.
Letting
\begin{eqnarray}
\pi_T(\thetab) &=& \Nc(\thetab; \mub_T,\Sigmab_T) , \\
\pi_S( \thetab \vert \beta) &=& \Nc(\thetab; \mub_S,(1/\beta)\Sigmab_S) ,
\end{eqnarray}
the resulting tempered Gaussian posterior is defined by the first two moments
\begin{eqnarray}
\Sigmab^{-1}_p(\beta) &=& \Sigmab_T^{-1} + \beta \Sigmab_S^{-1} \\
\mub_p(\beta)         &=& \Sigmab_p(\beta) \left[ \Sigmab_T^{-1} \mub_T +  \beta \Sigmab_S^{-1} \mub_S  \right]
\end{eqnarray}
Note that for Gaussian distributions, power tempering preserves the mean of the source likelihood while uniformly scaling the covariance.
This scaling preserves the correlation coefficient between variables and so maintains the same correlation structure while diffusing knowledge by decreasing the overall uncertainty as $\beta$ goes to zero.
This can be seen by noting that the correlation matrix $\Rb$ (matrix of correlation coefficients) can be extracted from the covariance matrix $\Sigmab$ as $\Rb = \Db^{-1} \Sigmab \Db^{-1} $ where $\Db = \sqrt{{\rm diag} (\Sigmab)}$.
Hence, it can be easily shown that the correlation coefficients remain unchanged when scaling the covariance matrix by $\beta$ as is done in power tempering.
While we focus on power tempering in this work, note that other tempering transformations, such as ones based on convolution, could be adopted in the proposed work flow resulting in different mechanisms of diffusing the knowledge encoded in the likelihood PDFs. The optimal choice of tempering transformation might be dependent on the data distribution and the models being calibrated.

\subsection{Objective functions} \label{sec:objectives}

In this section, we describe four objective functions considered for the empirical Bayes formulation of transfer learning presented in \sref{sec:transfer_learning}.
These objectives depend on the tempering parameter $\beta$ through the source likelihood and/or target posterior.
The objective functions are chosen to be sensitive to common features between the source and target likelihoods and behave in a qualitatively similar fashion.
The resulting optimal tempering parameter $\optbeta$ determines how much knowledge from the source task should be diffused in the process of transferring information to the target task.

The first objective is based on the Expected Data Fit (EDF), an information-theoretic quantity defined by $\expectation_{\pi_p}\left[ \log \pi_T \right]$ that effectively measures the average data fit provided by the posterior distribution $\pi_p$ over parameters.
It is commonly used in Bayesian inference and statistical modeling to measure how well a probabilistic model fits observed data.
It quantifies the likelihood of observing the given data under the model \cite{bishop2006pattern}.
Hence, by maximizing the EDF, we improve the target data fit which leads to the following objective function and equivalent minimization problem
\begin{equation}
\argmin_\beta - \expectation_{\pi_p( \thetab \vert \beta)} \left[ \log \pi_T(\thetab) \right]
\label{eq:EDF}
\end{equation}
This problem can be expressed in several different equivalent forms including:
\begin{align}
\argmin_\beta - \expectation_{\pi_p( \thetab \vert \beta)} \left[ \log \pi_T(\thetab) \right]
&= \argmin_\beta C(\pi_p( \thetab \vert \beta), \pi_T(\thetab)) \label{eqn:edf_kld_form} \\
&= \argmin_\beta  \kld{\pi_p( \thetab \vert \beta)}{\pi_T(\thetab)} + H(\pi_p( \thetab \vert \beta)) \nonumber  \\
&= \argmin_\beta  \kld{\pi_p( \thetab \vert \beta)}{\pi_S( \thetab \vert \beta)} - e(\beta) \nonumber
\end{align}
where $C$ denotes the cross-entropy, $D_\text{KL}$ the Kullback–Leibler divergence, $H$ the entropy, and $e = \int \pi_T(\thetab) \pi_T^0( \thetab \vert \beta) \mathrm{d}\thetab$ the model evidence.
For example, minimizing the negative EDF is seen to be equivalent to minimize the cross entropy between the tempered posterior $\pi_p( \thetab \vert \beta)$ and target likelihood $\pi_T(\thetab)$, a common objective function in machine learning applications.

In the case where all distributions are Gaussian, the EDF takes the form
\begin{equation}
\begin{split}
\argmin_\beta \frac{1}{2} \bigl[ (\mub_p(\beta) - \mub_T)^T \Sigmab_T^{-1}	(\mub_p(\beta) - \mub_T) \\ +\tr\left( \Sigmab_T^{-1}\Sigmab_p(\beta)  \right) + \log \det{2 \pi \Sigmab_T}  \bigr]
\end{split}
\end{equation}

The second objective is based on the  KL-divergence which also originates from information theory and statistics \cite{kullback1951information}.
It quantifies the difference between two probability distributions and is widely used in various fields, including machine learning, to measure the dissimilarity between two probability distributions.
It plays a pivotal role in tasks such as model comparison, optimization, and information retrieval. By taking the KL-divergence term from \eref{eqn:edf_kld_form}, we arrive at a second objective function
\begin{equation}
\argmin_\beta \kld{\pi_p( \thetab \vert \beta)}{\pi_S( \thetab \vert \beta)}
\end{equation}
which we denote as KLD.

For Gaussian distributions, the optimization has the form
\begin{equation}
\begin{split}
\argmin_\beta \frac{1}{2} \bigl[(\mub_p(\beta) - \mub_S)^T\beta \Sigmab_S(\beta)^{-1} (\mub_p(\beta) - \mub_S)\\ + \tr \left(  \beta \Sigmab_S^{-1}\Sigmab_p(\beta)\right)  - \log \frac{\det{\Sigmab_p(\beta)}}{\det{(1 / \beta)\Sigmab_S(\beta)}}  - d \bigr]
\end{split}
\end{equation}
Recall that $d$ is the degree of the PCE model.

The third objective is based on the model evidence $e =\int \pi_T\pi_S(\beta) \d{\thetab}$ which forms the posterior normalizing factor of the distribution $\pi_S(\beta) \pi_T \propto \pi_p(\beta)$.
It also known as the marginal likelihood, in Bayesian statistics.
In Bayesian inference, the model evidence provides the probability of the observed data marginalized over all possible model parameters $\thetab$, in essence giving the likelihood that a particular model resulted in the data \cite{mackay2003information}.
Hence, this quantity is often used in model selection, a Bayesian strategy where the most appropriate model is selected from a family of possible models.
By maximizing this quantity, we arrive at our third objective function
\begin{equation}
\argmax_\beta  \int \pi_T(\thetab)\pi_S( \thetab \vert \beta) \d{\thetab}
\end{equation}
which we denote by ME.

For Gaussian distributions, this reduces to
\begin{equation}
\argmax_\beta  \hspace{1mm} \mathcal{N}(\mub_T;\mub_S,\Sigmab_T+ (1/ \beta)\Sigmab_S)
\end{equation}

The fourth and final objective, Dice Similarity (DS) is widely used in image segmentation and pattern recognition  \cite{zijdenbos1994morphometric} to measure the similarity between two sets or regions. It has also been used to compare probability distributions \cite{lu1989multivariate} where it can be seen as a normalized inner product that reflects how much the two distributions overlap.
The DS satisfies $0 \leq D_{\text{dice}} (p,q) \leq 1$ for distributions $p,q$ with $D_{\text{dice}} (p,q) = 1$ if and only if $p=q$.
Hence, this provides an interpretable, normalized notion of similarity between two distributions that can be use to formulate the objective
\begin{equation}
\argmax_\beta  D_\text{dice}(\pi_S( \thetab \vert \beta),\pi_T(\thetab))
\end{equation}
where we omit the lengthy Gaussian formula for the sake of brevity.

Note that objective function from each of these four optimization problems can be expressed as $\int \pi_T(\thetab) \pi_S( \thetab \vert \beta) w( \thetab \vert \beta) \d{\thetab}$ where $w$ is a, possibly $\beta$-dependent, weighting function.
Namely, each is a weighted inner product between the target and source likelihoods and measures some form of overlap between the two distributions.
These inner products are also associated with norms if the two functions are normalized.

\section{Results} \label{sec:numerical_results}

In this section we first explore the behavior of four objective functions introduced in \sref{sec:methodology} using easily interpretable test problems.
\sref{sec:domain_adaptation} details an investigation of transfer learning for \emph{domain} adaptation  across all four objective functions.
We compare the performance of the optimal transfer evoked by each objective to no transfer and full transfer for four domain overlaps with three different polynomial models.
This exhaustive study is used to down-select from the set of objective functions to a single, best-performing one.
Subsequently, in \sref{sec:ishigami}, we apply the proposed method to a \emph{task} adaptation problem using a more complex generative model to validate previous observations and show that they hold for task adaptation.
Finally, in \sref{sec:subsurface}, the transfer learning framework is deployed on a problem of pragmatic interest involving PCE surrogates calibrated to models of subsurface scattering.
Here, again we compare to no and full transfer to demonstrate the method outperforms each.

\subsection{Domain adaptation}
\label{sec:domain_adaptation}

Domain adaptation addresses the challenge of training models on source data and generalizing their performance to target data with different distributions.
While traditional calibration assumes that source and target data come from the same distribution, real-world scenarios often involve different feature spaces or evolving statistical properties.
Collecting target data that perfectly matches the source distribution can be challenging, and utilizing related annotated data is often preferred to creating new labeled datasets.
Domain adaptation minimizes distribution differences between the source and target domains to enable effective knowledge transfer.

In each of the examples in this section, we utilize the pushed-forward posterior distribution (PFP) to evaluate the target task model after transfer learning.
The PFP leverages a parameter-to-prediction mapping that establishes a connection between parameters and the observed outcomes  \cite{Butler-etal-2018}.
The PFP is the pushforward density \cite{bogachev2007measure, papamakarios2021normalizing} of the posterior distribution (\ref{eq:mod_bayes}) over model parameters $\thetab$ through the model function $f(x;\thetab)$ at unobserved data $x = x^i$.
In the case where the tempered posterior is Gaussian $\pi_p(\thetab \vert \beta) = \Nc(\thetab \vert \mub_p(\beta),\Sigmab_p(\beta))$, the PFP density is defined by
\begin{equation}
\operatorname{PFP}(\hat{y}^i \vert \beta) = \Nc(\hat{y}^i \vert \mub(\beta), \Sigmab(\beta))
\label{eq:pfp}
\end{equation}
where $\mub = \Ab(x^i) \mub_p(\beta)$ and $ \Ab(x^i) \Sigmab_p(\beta) \Ab^T(x^i)$,
which follows from the standard results on linear transformations of a Gaussian random variables \cite{petersen2008matrix} for model that is linear in its parameters such as PCE.
Recall $\Ab$ is the Vandermonde matrix.
By computing the log probability of the PFP across data $(x_i,\hat{y}_i)$, $i=1,\ldots,m$ as a function of $\beta$
\begin{equation}
\operatorname{LPFP}(\beta) = \sum_{i=1}^m \log \operatorname{PFP}(\hat{y}^i \vert \beta)
\label{eq:sum_pfp}
\end{equation}
we obtain a metric that helps in assessing the model quality, aggregating uncertainties effectively, and making informed decisions based on model performance on unobserved data.

\subsubsection{Comparison of objective functions}
In the first demonstration, we aim to transfer information between the source and target domains by leveraging the knowledge learned from source domain to improve the performance of a model in the target domain when there is a shift or mismatch between the two domains.
An good objective function will perform better than no or full transfer of knowledge and avoid negative learning relative to these two baselines.

For this demonstraton, we focus on the construction of PCE-based surrogates for a (ground truth) one-dimensional third-order polynomial:
\begin{equation}
y = 0.3 ((1/3) x^3 - 2.25 x).
\label{eq:ytruth}
\end{equation}
The calibration problem is constructing a Legendre polynomial-based surrogate for the source domain $x \in [-0.2 , 0.3]$  and transferring  the learned knowledge to various target domains of length $0.4$ in the interval $[-3.0, 3.0]$.
For each test the interval encompassing the source and target domains is scaled and shifted using linear transformations to fit within the range $[-1, 1]$.
We approximate the data in the source and target domains using a truncated PCE:
\begin{equation}
\begin{aligned}
\hat{y}^i = \hat{y}(x^i) \approxeq f(x^i; \thetab) = \sum_{j = 0}^{p}  \theta_j \psi_j(x^i) ,
\end{aligned}
\label{eq:legendre-pce}
\end{equation}
where $\psi_j(x^i)$ is the orthonormal Legendre polynomial of degree $d\in \{1,2,3\}$ and $\thetab$ is the vector of model parameters.
The PCE of degree $d$ has $p=d+1$ parameters in this 1D case.
To mimic noisy observations, we add random Gaussian white noise  as
\begin{equation}
\hat{y}^i = f(x^i; \thetab) + \epsilon_i.
\label{eq:noisy-legendre-pce}
\end{equation}
and uniformly sample 16 data points from source domain and 4 data points from chosen target domain.
We form the  $\Ab_s \in \Rbb^{16 \times p}$ and $\Ab_t \in \Rbb^{4 \times p}$  Vandermonde matrices of orthonormal Legendre polynomial basis functions to calculate the prior and likelihood PDFs.
The optimal tempering parameter $\optbeta$ is obtained using objective functions EDF, KLD, ME, and DS from \sref{sec:objectives}.

\begin{figure}[!htb]
\centering
\includegraphics[width=\textwidth]{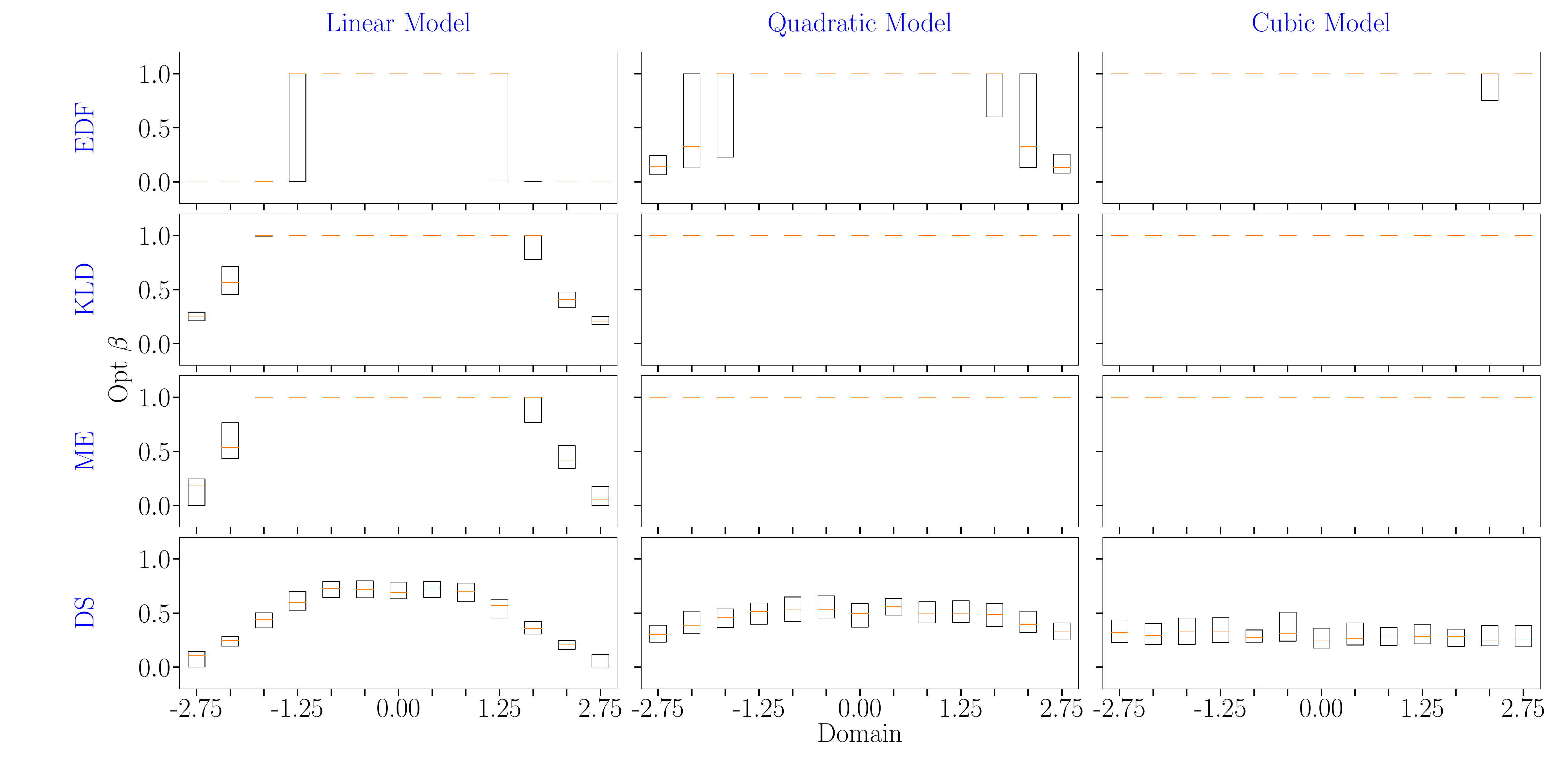  }
\caption{Optimal tempering parameter $\optbeta$ for various objective functions and model types across different target domains.}
\label{fig:sec_da_optbeta}
\end{figure}

\fref{fig:sec_da_optbeta} shows box plots of an ensemble of 100 realizations which illustrate varying levels of knowledge transfer across different objective functions and model types.
In the figure, the boxes represents the interquartile range ($IQR$), showing the middle 50\% of the data, while the line within the box indicates the median.
In \fref{fig:sec_da_optbeta}, the $x$-axis indicates the distance between the mean of source and the mean of target domains, while the  $y$-axis represents the optimal amount of knowledge transferred from the source to the target domain, as measured by the tempering exponent $\beta$.
The value of $\beta$ ranges from 0 (indicating no transfer of knowledge) to 1 (representing full transfer), with intermediate values reflecting partial knowledge transfer.

The figure  reveals distinct patterns for different objective functions and model types.
Specifically, for quadratic and cubic models, the objective functions KLD and ME facilitate complete knowledge transfer from the source to the target domain.
However, for the linear model, the knowledge transfer is partial, especially when the target domain is far from the source domain, which appears to be optimal given the dissimilarity of the tasks.
On the contrary, when the target domain coincides or stays close to the source domain, the linear model transfers almost full knowledge.
In contrast, when using the DS objective function, the transfer of information is minimal for quadratic and cubic models.
For the linear model, it transfers more knowledge when the source and target domains coincide, but this knowledge transfer diminishes as the target domain moves away from the source domain.
Lastly, the EDF objective function results in complete knowledge transfer from the source to the target domain when the model is cubic.
For quadratic and linear models, complete knowledge transfer occurs when the source and target domains coincide, intersect, or remain close to each other, but no knowledge transfer takes place when the target domain moves away from the source domain.
The results portrayed in \fref{fig:sec_da_optbeta} demonstrate that the EDF objective function optimizes knowledge transfer appropriately and effectively between the source and target domains across different models types and domain configurations.

\begin{figure}[!htb]
\centering
\includegraphics[width=\textwidth]{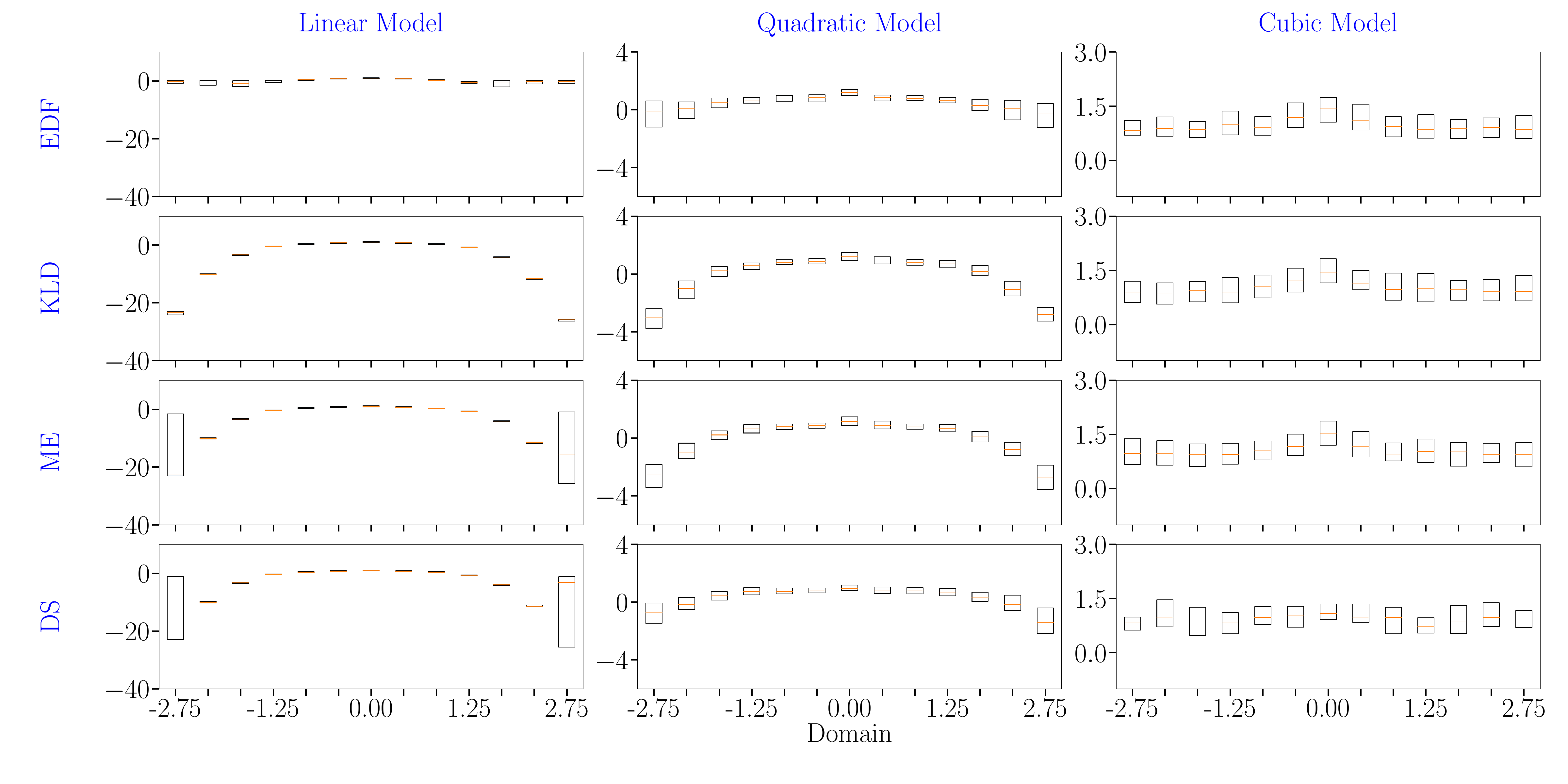  }
\caption{Performance gain in the proposed transfer learning against the no transfer baseline for various objective functions, model types and target domains; $y$-axis = performance gain against no transfer =  $\operatorname{LPFP}(\optbeta) - \operatorname{LPFP}(0)$. }
\label{fig:sec_da_diffPFP_B_B0}
\end{figure}

\begin{figure}[!htb]
\centering
\includegraphics[width=\textwidth]{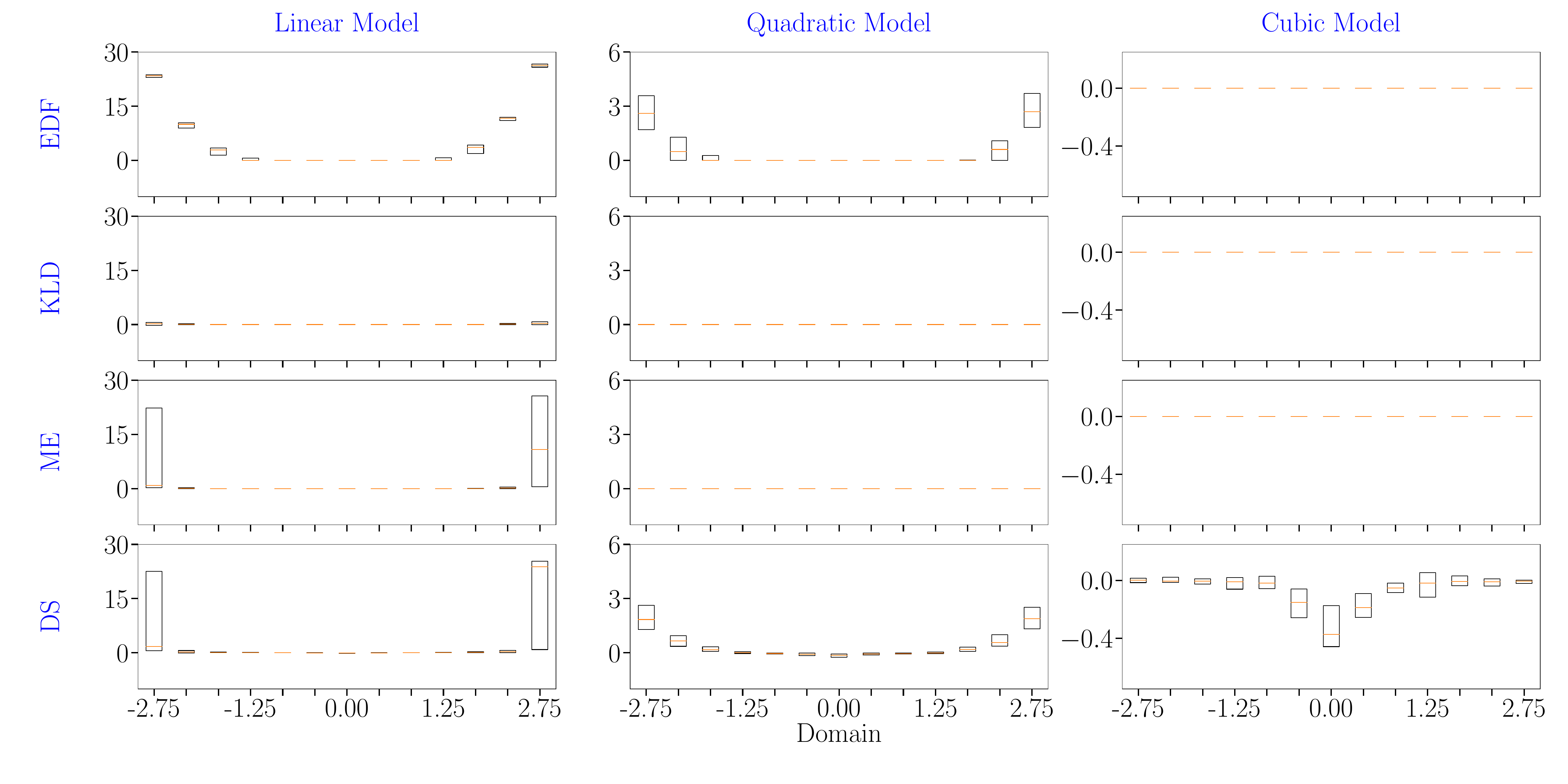  }
\caption{Performance gain in the proposed transfer learning against the full transfer baseline for various objective functions, model types and target domains; $y$-axis = performance gain against full transfer =  $\operatorname{LPFP}(\optbeta) - \operatorname{LPFP}(1)$. }
\label{fig:sec_da_diffPFP_B_B1}
\end{figure}

Next we calculate the LPFP for the same models and objectives .
Figures  \ref{fig:sec_da_diffPFP_B_B0}  and \ref{fig:sec_da_diffPFP_B_B1}  present a comparison of the LPFP scores for optimal transfer $\beta = \optbeta$ relative to  no transfer $\beta = 0$ and to full transfer $\beta = 1$ , respectively.
The $x$-axis in each panel is the domain shift, and the $y$-axis is the difference in LPFP.
Where the difference is positive this indices the optimal transfer, as defined by the selected objective, has outperformed the respective baseline.
Across all the scenarios, it appears that only EDF generally avoids negative transfer and has improvements over both baselines.
The KLD objective has some regions of comparable performance but also significant negative learning.
It is also noteworthy that the performance relative to no transfer is typically the complement of that relative to full transfer.

Figure \ref{fig:sec_da_diffPFP_B_B0}  illustrates that regardless of the chosen objective function, the $\optbeta$ consistently outperforms the no transfer approach for the cubic model.
Similarly, for the quadratic and linear models, $\optbeta$ exhibits superior performance as long as the target domain remains close to the source domain, irrespective of the objective function used.
However, when the target domain moves significantly away from the source domain, the performance of $\optbeta$ diminishes for the quadratic model, especially for KLD and ME objectives.
Additionally, for the linear model, no transfer surpasses the performance of $\optbeta$ by a substantial margin for KLD, ME, and DS objective functions, whereas $\optbeta$ performs similarly to no transfer for the EDF objective.

Figure  \ref{fig:sec_da_diffPFP_B_B1}  illustrates that in the case of the cubic model, both the $\optbeta$ and $\beta = 1$ yield comparable results for the EDF, KLD, and ME objective functions. However, for the DS objective function, when the target domain intersects, merges, or remains closely aligned with the source domain, full transfer outperforms optimal transfer.
Moving on to the quadratic model, once again, the scenarios where $\optbeta$ and $\beta = 1$ perform equally are observed with the KLD and ME objective functions.
Nevertheless, when the target domain deviates significantly from the source domain, $\optbeta$ exhibits superior performance for the EDF and DS objective functions.
For the linear model, an interesting pattern emerges. In cases where the target domain lacks intersection with the source domain, $\optbeta$ significantly outperforms full transfer ($\beta = 1$) in the context of the EDF objective function. However, across the KLD, ME, and DS objective functions, $\optbeta$ consistently outperforms $\beta = 1$ when the target domain moves far away from the source domain.

It is important to emphasize that the true underlying model is cubic, as represented by \eref{eq:ytruth}. When approximating this truth using truncated PCE, whether through a linear or quadratic model (\eref{eq:legendre-pce}  with $d = 1$ or $2$), no knowledge transfer is expected when there is insignificant similarity between the trends in the target and source domains.
On the other hand, transfer, whether partial or full, occurs when an intersection between the two domains exists, and the extent of transfer is contingent upon the degree of intersection.
Figures  \ref{fig:sec_da_optbeta}, \ref{fig:sec_da_diffPFP_B_B0}  and \ref{fig:sec_da_diffPFP_B_B1} provide clear evidence that the EDF objective function consistently outperforms other objective functions (KLD, ME, DS) across all model types (linear, quadratic, and cubic) when used with the $\optbeta$ strategy.
These findings underscore the efficacy of the $\optbeta$ strategy and the prominence of the EDF objective function in the context of domain adaptation scenarios.
Hence, for the further analysis in this paper, we consider only EDF objective function.

\subsubsection{Illustration of performance of the EDF objective}

\begin{figure}[!htb]
\centering
\includegraphics[width=0.6\textwidth]{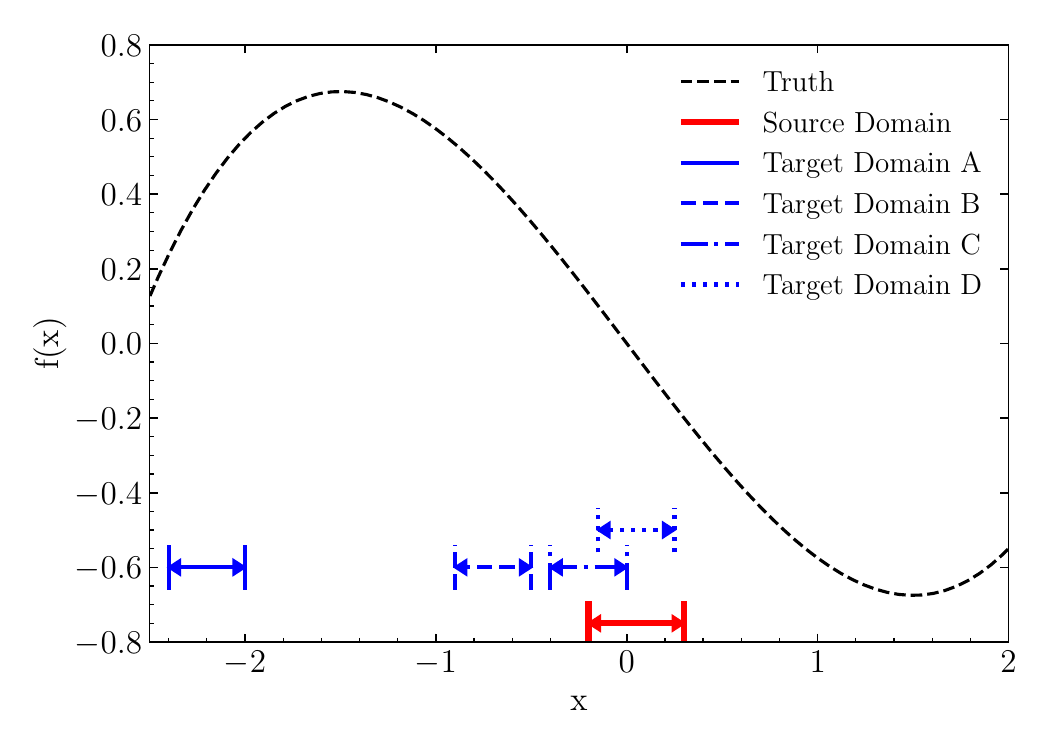  }
\caption{Representation of truth, source and target domains.}
\label{fig:sec_da_rep}
\end{figure}

Now we demonstrate the performance of EDF objective function for a given model type, target domain, and three different $\beta$ values, namely, $\beta = 0$ (no transfer), $\beta = 1$ (full transfer) and $\beta =  \optbeta$ (optimum transfer).
Figure \ref{fig:sec_da_rep} shows the underlying truth given by \eref{eq:ytruth}, source domain and four different target domains denoted by A (target domain is far from the source domain), B (target domain is close to the source domain), C (target domain intersects partially with the source domain), and D (target domain merges with the source domain).
In order to understand the performance of all models (linear, quadratic and cubic) over various target domains in transfer of knowledge, we calculate the PFP to obtain a probabilistic predictions for each target domain (A, B, C, and D). We randomly generate 16 data points from source domain and 4 data points from target domain for 100 different realizations using Latin Hypercube Sampling.

\begin{figure}[!htb]
\centering
\includegraphics[width=\textwidth]{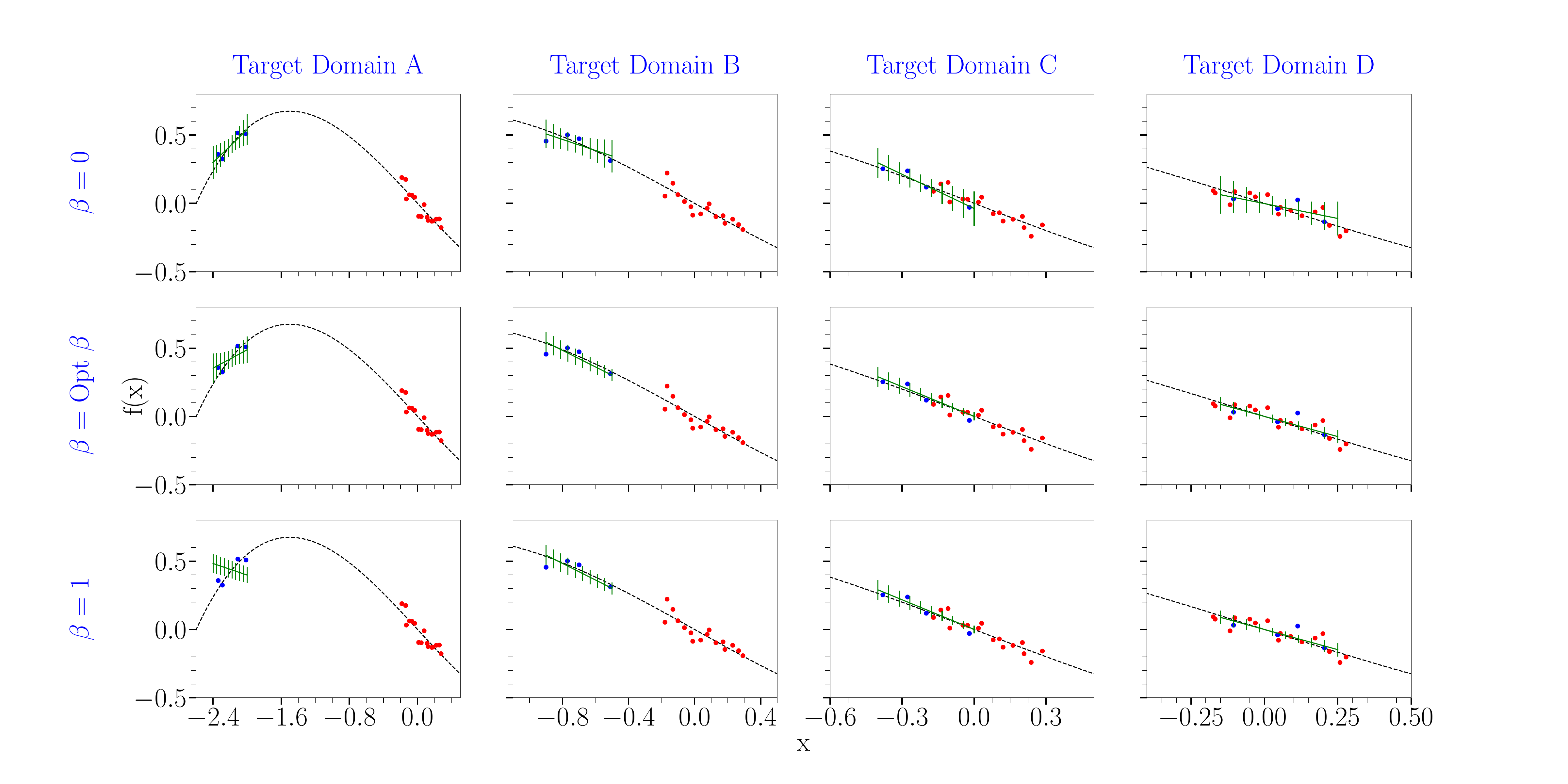  }
\caption{PFP for the linear model with fixed source domain and four different target domains for EDF objective. Red: source data, blue: target data, green: model fits.}
\label{fig:sec_da_linearPFP}
\end{figure}

Figure \ref{fig:sec_da_linearPFP}  shows the PFP of the linear model, where the source domain remains constant, while varying target domains are considered.
Each plot includes a 95\% confidence interval derived from 100 instances.
For target domain A, it is notable that the absence of knowledge transfer yields a closer fit to the true model compared to optimal knowledge transfer and full knowledge transfer.
However, the trend shifts for target domains B, C, and D.
In these cases, optimal knowledge transfer and full knowledge transfer provide a markedly better fit to the underlying truth while exhibiting reduced uncertainty, in contrast to the case of no knowledge transfer.
Applying this same analysis to a quadratic model fitted to data gave similar treends, which have been omitted for brevity.

\begin{figure}[!htb]
\centering
\includegraphics[width=\textwidth]{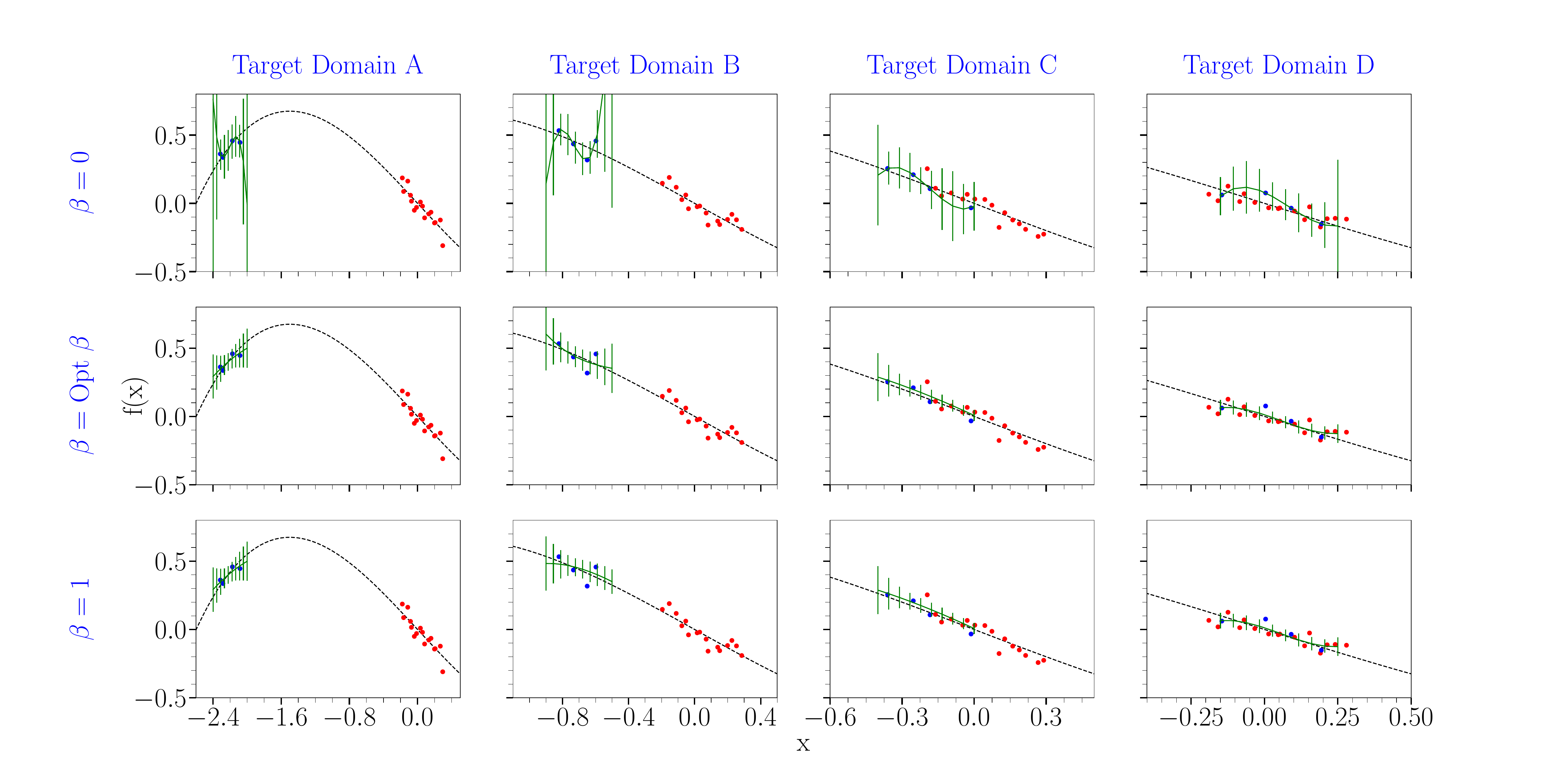  }
\caption{PFP for the cubic model with fixed source domain and four different target domains for EDF objective. Red: source data, blue: target data, green: model fits.}
\label{fig:sec_da_cubicPFP}
\end{figure}

\fref{fig:sec_da_cubicPFP}  depicts the cubic model fits, adhering to a consistent source domain while experimenting with diverse target domains.
Similar to the previous models, each plot is accompanied by a 95\% confidence interval derived from 100 instances.
Unlike the linear and quadratic models, in the case of the cubic model, optimal knowledge transfer and full knowledge transfer distinctly outperform the absence of knowledge transfer.
This improved performance holds true across all target domains (A-D), with both optimal and full knowledge transfer demonstrating a superior alignment with the underlying truth while exhibiting reduced uncertainty.

Following a comprehensive investigation into transfer learning for domain adaptation encompassing four distinct objective functions across three models (linear, quadratic and cubic), our findings indicate that:
(a) the EDF objective function optimizes knowledge transfer effectively between the source and target domains across different models types and domain configurations;
(b) the other objective functions (KLD, ME \& DS) exhibit similar performance to EDF across various domains, only when the model is generative;
(c) the performance gain in the proposed transfer learning ($\beta = \optbeta$) with the EDF objective outperforms no transfer ($\beta = 0$) and full transfer ($\beta = 1$) across various domains and models considered.
Consequently, we have elected to adopt the EDF objective function as our primary approach for illustrating task adaptation in the following section.

\subsection{Task adaptation with the Ishigami function}
\label{sec:ishigami}
In contrast to domain adaptation, which we explored in the previous section, task adaptation considers the case where the source and target data come from the same distributions but the forward models, or tasks, themselves are different.
We assume prior knowledge that the two tasks, while not exactly the same, share similarities that can be exploited to transfer knowledge from source to target.

\begin{figure}[htp!]
\centering
\includegraphics[width=0.9\textwidth]{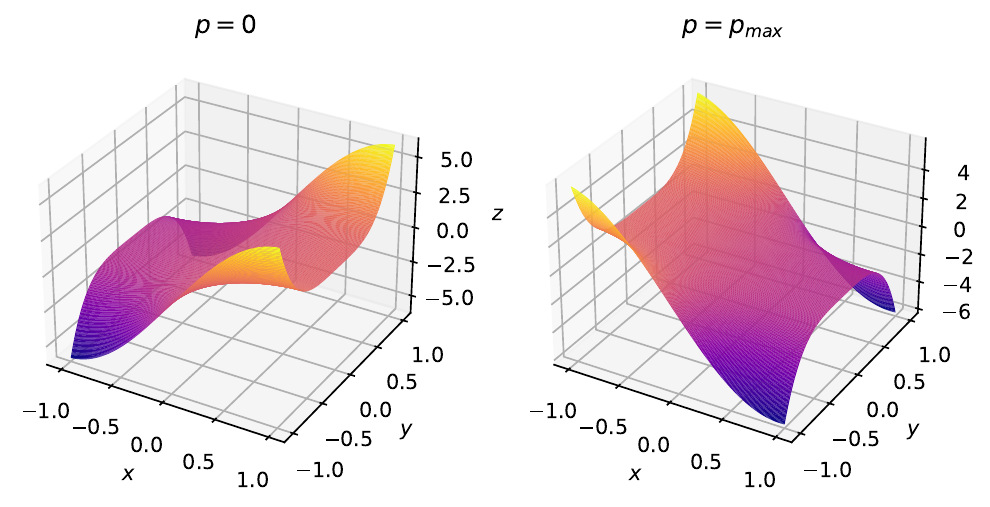}
\caption{Ishigami function at minimum and maximum $p$ values.}
\label{fig:ishigami_3d}
\end{figure}

In addition, we would like to illustrate how the proposed method  performs on a higher-dimensional generative model that is not polynomial.
Hence, we construct a calibration task based on the Ishigami function \cite{ishigami1990importance} defined as $f(x_1,x_2,x_3; \thetab) = \sin(x_1) + a \sin^2(x_2) + b x_3^4 \sin(x_1)$, where $\thetab = (a,b)$.
The Ishigami function exhibits strong nonlinearity and nonmonotonicity as well as complex dependence on $x_3$  \cite{sobol1999use}.
For simplicity, we take $a=0$, $b=1$ so that we have a function of two variables $x_1$, $x_3$ which we denote as $f(x,y)$.
We introduce parametric dependence on a single parameter $\theta$ by translating $x$, i.e., $f(x,y;\theta)=f(x-\theta,y)$.
The source and target domains are then fixed to be $[-1,1]^2$ while different tasks are obtained by varying $\theta$.
Hence, the generative model is given by
\begin{equation}
f(x,y; \theta) = \sin(x-\theta) +  y^4 \sin(x)
\label{eq:ishigami}
\end{equation}
We let $p = || \theta_T- \theta_S ||$ represent the difference between the $\theta_T$, $\theta_S$, the source and target values of $\theta$, respectively and display $f(x,y;\theta)$ at the minimum and maximum values of $p$ in \fref{fig:ishigami_3d}.

The PCE model has the form of a tensor-product polynomial basis over $\Rbb^2$
using the Legendre polynomials with maximum degree $d=3$.
Here, the number of PCE coefficients gives $\thetab \in \Rbb^{10}$.
To carry out transfer learning, we focus on the EDF objective function \eqref{eq:EDF} to determine $\beta$, as it showed superior behavior in domain adaptation in \sref{sec:domain_adaptation}.
By looking at the optimal $\beta$ values and performance metrics such as pushforward error, we can examine the performance of transfer learning in an analogous manner to \sref{sec:domain_adaptation}.

\begin{figure}[htp!]
\centering
\includegraphics[width=0.8\textwidth]{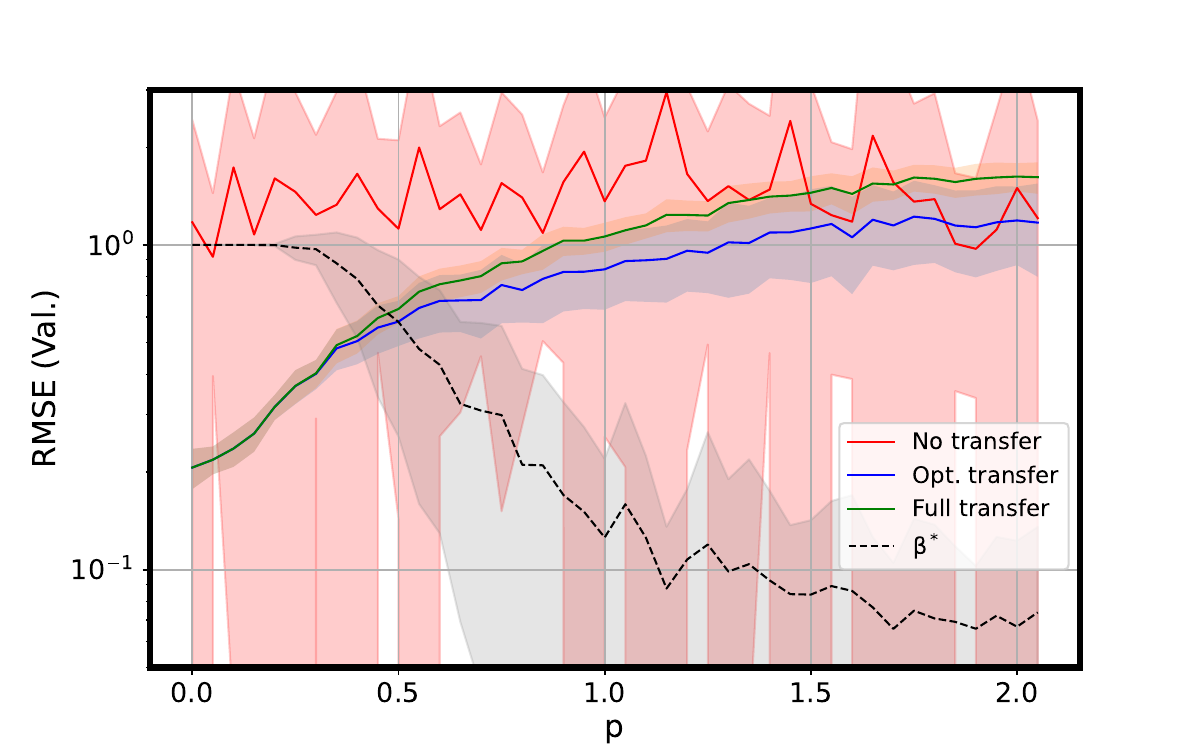}
\caption{Ishigami problem: RMSE of target model evaluated with the mean parameters of the PFP distribution.
The cases of no transfer, optimal transfer, and full transfer are displayed in red, blue, and green, respectively.
Optimal $\beta$ is shown in dashed black.
Each quantity is plotted as a function of $p$.
The confidence bounds reflect the variance from random samplings of the source and target domains and are drawn at 1 standard deviation.}
\label{fig:ishigami}
\end{figure}

To measure the predictive performance of the source and target PCE surrogate models, a held-out set of  $N_{\text{val}} = 1000$ source and target data was used to define a validation root mean squared error (RMSE) for the source and target models using the mean parameters from the PFP distributions.
This approach is motivated by the ability of surrogate models, whether PCE or others, to exactly fit or \textit{overfit} datasets that are sparse with respect to the number of model parameters.
Hence, we expect the target model to fit the data exactly so that the RMSE over the data used to calibrate the model is uninformative.
To account for the effect of sampling on the transfer learning methodology, we average the RMSE performance over $N = 500$ trials where data is randomly sampled from the source and target domains.
The number of source data points is $N_S = 40$ while the number of target data points is $N_T = 11$.
This is just sufficient to obtain an over-determined least squares problem for the $\thetab$.

\fref{fig:ishigami} shows the mean validation RMSE and confidence intervals for optimal transfer as well as optimal $\beta$ values as a function of the difference $p$ between the source and target values for $t$.
Also displayed are the cases of no transfer, $\beta = 0$, and full transfer $\beta=1$ to provide baseline comparisons.
Similar to \fref{fig:sec_da_optbeta}, we see that in the top of \fref{fig:ishigami}, the optimal $\beta$ tends to zero as $p$ increases or, equivalently, as the source and target task become more distinct.
At $p=0$, the source and target task problems coincide yielding an average $\beta$ with low uncertainty.
Hence, the regime where $p$ is sufficiently small, we see that full transfer and optimal transfer perform similarly while no transfer has significantly large RMSE.
As $p$ increases, the accuracy of full transfer decreases and ultimately becomes worse than no transfer near $p=2$.
Observe that optimal transfer adapts to the difference between the source and target tasks and subsequently outperforms no transfer and full transfer over the entire range of $p$.
Because of the sparsity of the target dataset, we see large uncertainty in the RMSE for the no transfer case as well as the optimal $\beta$ values.

\subsection{Domain adaptation with PCE models of subsurface scattering}
\label{sec:subsurface}

\begin{figure}[!ht]
\centering
\includegraphics[width=0.7\textwidth]{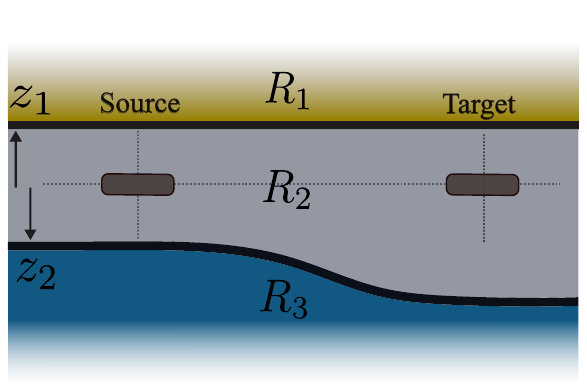}
\caption{Schematic diagram of the 3-layer subsurface transfer learning problem with physical model parameters given by the layer resistivities $R_1$, $R_2$, $R_3$ and boundaries $z_1, z_2$.
The target task is differentiated from the source task by varying the depth $z_2$ of the second boundary or by changing the resistivity of the third layer $R_3$.
This represents the geosteering scenario where rock layers have the same composition by spatially varying thicknesses.}
\label{fig:subsurface_layers}
\end{figure}

In this practical demonstration, we apply the probabilistic transfer learning methodology to Bayesian inversion of a 1-dimensional subsurface scattering model over differing domains.
Problems such as this arise in applications such as geosteering for drilling and resource extraction \cite{lesso1996principles} where determining properties of strata is paramount.
The forward model considers a collection of $n_l$ subsurface layers depicted in the schematic \fref{fig:subsurface_layers} under the simplifying assumption that physical properties vary depthwise ($z$) in layers.
The quantities of interest are the electrical resistivities $R_i; i=1,\ldots,n_l$ for each of the layers, measured in units of ohm-meter $\mathrm{\Omega} \cdot \mathrm{m}$, as well as the $z$-depth $z_j, j=1,\ldots, n_l -1$ of the boundary of each layer, measured in $\mathrm{ft}$.
A transmitter and receiver device is placed at the origin which we take to be $z=0$.

For the sake of interpretability, we consider a $3$-layer subsurface model with $5$ total parameters consisting of the resistivities $R_1$, $R_2$, $R_3$ of each layer along with two $z$-depths $z_1$, $z_2$ that define the location of the two layer boundaries relative to the sensor.
The generative model is then given by solving Maxwell's equations over the 1-dimensional domain.
We approximate the generative model with a PCE over the five parameters
of maximum degree $d=3$, yielding a total of $56$ coefficients so that $\thetab \in \Rbb^{56}$.
The transfer learning framework methodology is applied to a parametric sequence of domain adaptation problems defined by source and target PCE surrogate approximation of the subsurface model.
The source parameters ranges are kept fixed while the target domain is varied linearly by modifying the thickness of the second layer, which contains the sensor, or the resistivity range for the third layer.
This mimics a geosteering scenario where drilling has moved from one location to another and either the subsurface layer thickness or composition is spatially varying.

Similar to \sref{sec:domain_adaptation}, this setup allows us to directly measure the performance gain achieved by transfer learning as a function of the discrepancy between source and target data distributions.
The ranges of the source parameters are displayed in \tref{table:subsurface_params}.
The target domain is shifted in two ways.
First, the range for the boundary $z_2$ between layers 2 and 3 is shifted from $[1,2]$ to $[4,5]$ for a maximum difference of $3 \hspace{1mm} \mathrm{ft}$.
Second, $R_3$, the resistivity of the third layer, from $[7,9]$ to $[17,19]$, for a maximum difference of $10 \hspace{1mm} \mathrm{\Omega} \cdot \mathrm{m}$.
The source problem is calibrated on $N_S = 200$ data points uniformly sampled over the domain in \tref{table:subsurface_params} while the target is calibrated using only $N_T = 57$ data points.
This sample size is just sufficient to obtain an over-determined least squares problem for the $56$ PCE parameters.
As in \sref{sec:ishigami}, we use the RMSE over the set of validation data to measure the success of transfer learning.

\begin{table}[!ht]
\centering
\captionsetup{justification=centering}
\caption{Ranges of model parameters defining the source domain.}
\begin{tabular}{cccccc}
\toprule
Parameter & $R_1$ & $R_2$ & $R_3$ & $z_1$ & $z_2$ \\
\midrule
Unit &  $\mathrm{\Omega} \cdot \mathrm{m}$ &  $\mathrm{\Omega} \cdot \mathrm{m}$ &  $\mathrm{\Omega} \cdot \mathrm{m}$ & $\mathrm{ft}$ & $\mathrm{ft}$ \\
\midrule
min &  1 & 4  &  7 & -2 & 1   \\
max &  3  &  6  &  9  &  -1 & 2 \\
\bottomrule
\end{tabular}
\label{table:subsurface_params}
\end{table}

\begin{figure}[!ht]
\centering
\includegraphics[width=0.70\textwidth]{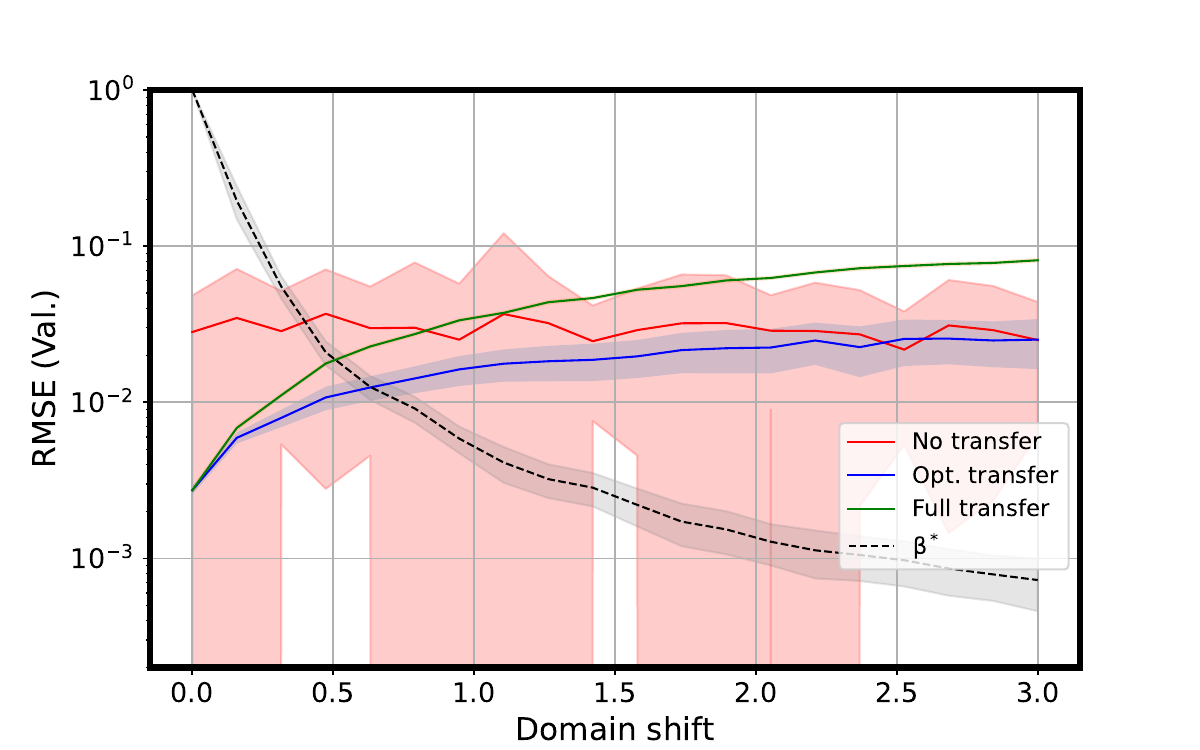}
\includegraphics[width=0.70\textwidth]{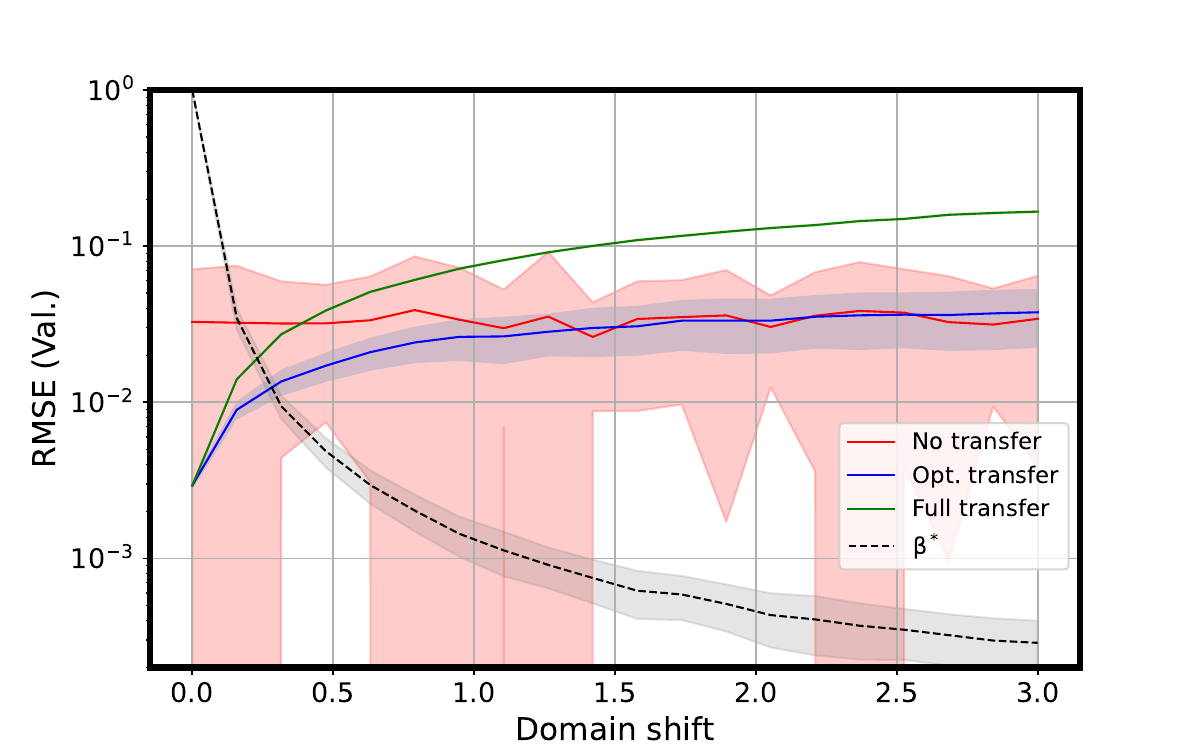}
\caption{Subsurface inversion: RMSE of target model evaluated with the mean parameters of the PFP distribution.
The cases of no transfer, optimal transfer, and full transfer are displayed in red, blue, and green, respectively.
Optimal $\beta$ is shown in dashed black.
Each quantity is plotted as a function of the shift between source and target domains.
The top displays the case where $z_2$ is shifted, while the bottom shows the behavior when $R_3$ is shifted.
The confidence bounds reflect the variance from random sampling of the source and target domains and are drawn at 1 standard deviation.}
\label{fig:subsurface_TL}
\end{figure}

The results for the shifting $z_2$ and $R_3$ are displayed in \fref{fig:subsurface_TL} where we see the validation error and its uncertainty for optimal transfer along with the two baseline cases of no transfer, and full transfer.
As seen in previous examples, the mean optimal $\beta$  decreases monotonically from one as the source and target domains become increasingly separated.
This shows that the probabilistic transfer learning methodology is able to diffuse source knowledge in response to large discrepancies between the two tasks.
When the two domains coincide, we see that no transfer results in a larger RMSE than both full and optimal transfer of knowledge which continue to behave similarly as long as the source and target domains remain close.
As the domains separate further, the two tasks become increasingly dissimilar, resulting in significant increase in the RMSE for full transfer while optimal transfer begins to adapt and coincide with no transfer.
Random sampling results in significant variance of the target model calibration.
Uncertainty in the no transfer RMSEs is considerably larger due to the sparsity of the target dataset.
Focusing on the mean values, we see that optimal transfer outperforms no transfer and full transfer over nearly the full sequence of shifted target domains.

\begin{figure}[!ht]
\centering
\includegraphics[width=0.70\textwidth]{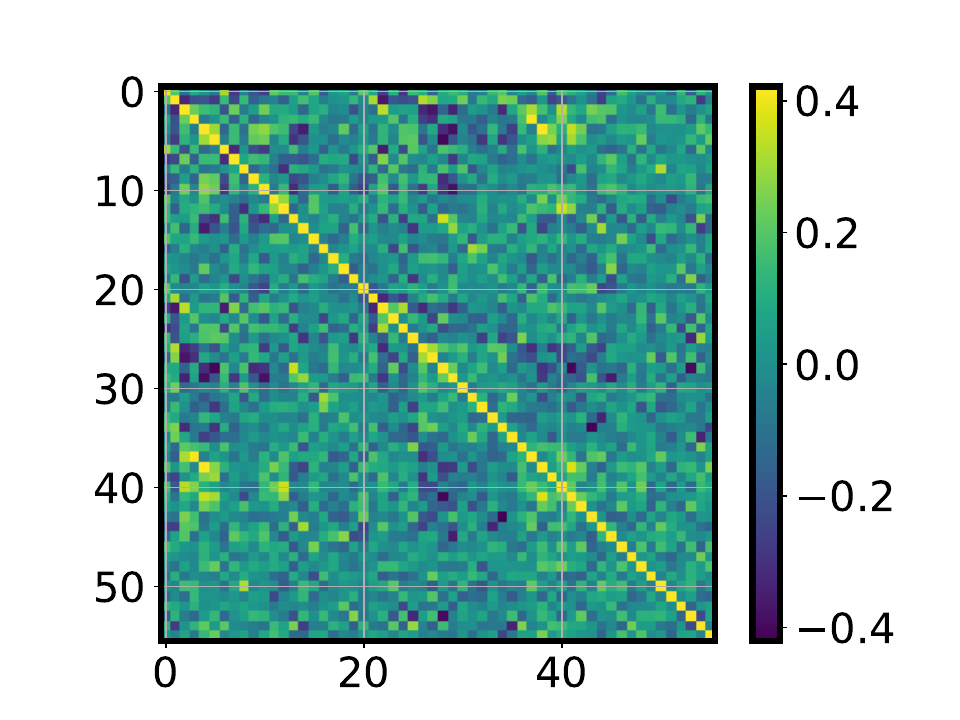}
\caption{Correlation coefficients of the source likelihood. The color range was set to span the minimum and maximum values of the off-diagonal entries as the diagonal entries are 1 by definition. Observe that using an orthonormal polynomial basis results in likelihood PDFs characterizing approximately uncorrelated unknown parameters.}
\label{fig:corr_coeff}
\end{figure}

A particular aspect of these results that merits further discussion is how the correlation structure of the source and target task affects the performance of transfer learning.
\fref{fig:corr_coeff} displays the matrix of correlation coefficients across all 56 PCE parameters of the source model after calibration using the source data.
Observe that the matrix displays strong diagonal dominance suggesting that the source posterior encodes Gaussian distributions for the parameters with weak correlations.
This results from using orthogonal polynomial basis functions for the PCE expansions which yield uncorrelated random variables.
This lack of correlation structure potentially uncouples and thus simplifies the transfer learning problem.
For example, recall that in \sref{sec:tempering}, it was noted how power tempering preserves the correlation structure since it uniformly scales the source likelihood's covariance matrix. Hence, while modifying the source likelihood's overall uncertainty, power tempering maintains the approximate independence across parameters during the transfer learning process.

\section{Conclusion} \label{sec:conclusion}
In this paper, we presented a probabilistic transfer learning framework based on informing a target task using optimal tempering of the Bayesian posterior of a source task.
Transfer learning is invaluable in scenarios with costly forward evaluations and reservoirs of related data.
It is also key in  data-driven modeling problems where sparse and/or noisy data makes model calibration ill-posed.
While transfer learning approaches exist for deterministic ML models, less focus has been given to uncertainty quantification.
The proposed method fills this need.

We extended transfer learning to Bayesian inference problems through an optimization-based methodology utilizing objective functions to control the diffusion of knowledge from source to target task.
We demonstrated that the method transitions between full and no transfer of information  while capturing uncertainty and avoiding negative transfer.
The method involved novel application of several existing information-theoretic and discrepancy metrics to construct suitable objective functions for optimally tempering the source task posterior distribution.
An extensive investigation of the behavior of transfer learning under these objective functions was carried out on a series of domain adaptation problems.
This was used to understand factors influencing the robustness of the methodology and to down-select to the best performing objective, namely the Expected Data Fit.
We showed that a significant performance gain is achievable using the transfer learning procedure on a task adaptation problem based on the Ishigami function.
Similar performance gains were seen on a geosteering application based on a 3-layer subsurface scattering model.
The optimal tempering procedure was seen to outperform naive full transfer, given by a standard Bayesian update, as well as ignoring the source data.

These results suggest that the proposed methodology can improve calibration performance in sparse and/or noisy data setting but more investigation is required to see whether this carries over to large-scale applications such as those arising in machine learning models.
Also investigation of more fine-grained control of the tempered correlation structure is left for future work.

\section*{Acknowledgements}
This work was supported by the U.S.  Department of Energy, Office of Science, SBIR/STTR Programs Office under Award Number DE-SC0021607.
It was also supported by the Laboratory Directed Research and Development program at Sandia National Laboratories, a multimission laboratory managed and operated by National Technology and Engineering Solutions of Sandia LLC, a wholly owned subsidiary of Honeywell International Inc.  for the U.S.  Department of Energy’s National Nuclear Security Administration under contract DE-NA0003525.
This report describes objective technical results and analysis.
Any subjective views or opinions that might be expressed in the report do not necessarily represent the views of the U.S.  Department of Energy or the United States Government.

\end{document}